\documentclass{article}

\PassOptionsToPackage{numbers, compress}{natbib}

\usepackage[main, final]{neurips_2026}

\usepackage[utf8]{inputenc} 
\usepackage[T1]{fontenc}    
\usepackage{hyperref}       
\usepackage{url}            
\usepackage{booktabs}       
\usepackage{amsfonts}       
\usepackage{nicefrac}       
\usepackage{microtype}      
\usepackage{xcolor}         
\usepackage{enumitem}
\usepackage{graphicx}
\usepackage{multirow}
\usepackage{makecell}
\usepackage{fontawesome5}

\usepackage{twemojis}
\usepackage{amsmath}
\usepackage{colortbl}
\usepackage{pifont}
\usepackage{soul}
\usepackage{arydshln}  
\usepackage{tcolorbox}
\tcbuselibrary{skins,breakable}
\usepackage{wrapfig}
\usepackage{float}
\definecolor{realsonyblue}{RGB}{0, 51, 102}
\definecolor{sonyblue}{RGB}{0, 0, 255}
\sethlcolor{sonyblue!15}
\newcommand{\hlsonyblue}[1]{{\sethlcolor{sonyblue!15}\hl{#1}}}
\newcommand{\hlpink}[1]{{\sethlcolor{pink!30!white}\hl{#1}}}

\newlist{tightitem}{itemize}{1}
\setlist[tightitem,1]{
  nosep,             
  leftmargin=8pt,    
  label=\textbullet  
}

\tcbset{
    myprompt/.style={
        colback=realsonyblue!10, 
        colframe=realsonyblue,   
        boxrule=1pt,         
        width=\textwidth,    
        fonttitle=\bfseries, 
        coltitle=white       
    }
}

\title{Omni-Interactive Universal Embedder}

\author{%
  Wei-Yao Wang\textsuperscript{\faGamepad}\thanks{Project Lead: Wei-yao.Wang@sony.com.}~, Kazuya Tateishi\textsuperscript{\faGamepad}, Shuyang Cui\textsuperscript{\faGamepad}, Christian Simon\textsuperscript{\faGamepad},\\ \textbf{Takashi Shibuya\textsuperscript{\faHeadphones}, Shusuke Takahashi\textsuperscript{\faGamepad}, Yuki Mitsufuji\textsuperscript{\faGamepad,\faHeadphones}} \\
  \textsuperscript{\faGamepad}Sony Group Corporation, \textsuperscript{\faHeadphones}Sony AI\\
}

\begin{document}

\maketitle

\begin{abstract}
Multimodal representation learning has been shifting from traditional two-tower architectures to large language model (LLM)-based embedders due to their strong instruction-following capabilities.
Despite this progress, existing approaches primarily focus on language and image modalities, which also remain the dominant modalities for user-conditioned interactions in current embedders.
In this paper, we propose the first Omni-Interactive Universal Embedder (OmniUE), which not only learns a unified embedding space across text, video, and audio by leveraging intermediate-layer representations from dedicated learnable tokens, but also supports omni-interactive querying, enabling users to provide inputs in the form of text, visual regions of interest, and audio spans.
Within OmniUE, visual and audio segmenters process diverse user interactions and integrate them with an omni-LLM to produce user-conditioned any-to-any embeddings via context aggregation.
To evaluate OmniUE's omni-interactive capabilities, we introduce OmniCHOIR, benchmarking models for omni-interactive compositional audio retrieval based on the given text, video, and audio as well as unimodal or multimodal interaction prompts.
OmniUE consistently surpasses state-of-the-art baselines across diverse modalities, with average improvements of 10.4\% on textual-interactive video benchmarks (MMEB-v2-video), 1.0\% on audio tasks (MAEB), 83.1\% on visual-interactive benchmarks (SCaR), and 21.9\% on our omni-interactive OmniCHOIR benchmark.
We believe that jointly advancing omni-modal representation learning and omni-interactive querying paves the way toward universal embedders.
Project page: https://wywywang.github.io/omniue-demo/.

\end{abstract}
\section{Introduction}

Multimodal embedding models have advanced from traditional two-tower-based models (e.g., \cite{DBLP:conf/icml/RadfordKHRGASAM21,DBLP:conf/icml/0001LXH22,DBLP:conf/iccv/ZhaiM0B23,DBLP:conf/eccv/WeiCCHZFRC24}) to multimodal LLM (MLLM)-based models due to their large-scale training corpora, support for arbitrary combinations of input modalities, and effective instruction-following abilities that allow users to extract corresponding embeddings from the same embedder using different text instructions.
This transition has been demonstrated to be effective in terms of reasoning and understanding over images \cite{DBLP:conf/iclr/JiangMYYZC25,DBLP:conf/cvpr/LiuZCJ0YWX25,DBLP:conf/cvpr/ZhangZXLDLXZLZ25,DBLP:journals/corr/abs-2511-19278,DBLP:conf/acl/0005W00ZWD25,DBLP:journals/corr/abs-2510-15543} and video \cite{DBLP:journals/corr/abs-2507-04590,DBLP:journals/corr/abs-2510-13515,DBLP:journals/corr/abs-2511-00405}; however, they only focus on text with the visual modality.
Recently, researchers have started exploring omni-modal embedders by finetuning omni-LLMs to encode heterogeneous inputs to a unified embedding space \cite{DBLP:journals/corr/abs-2601-03666,xiao2025scaling,DBLP:journals/corr/abs-2510-03458,DBLP:conf/sigir/MaGZZCL25}, which opens the way toward universal embedders, i.e., general-purpose embedders that support diverse modalities, tasks, and interaction scenarios within a unified framework.

Despite this progress toward universal embedders, current MLLM-based approaches mainly focus on incorporating new modalities with text or improving text-based instruction following, with text remaining the dominant medium for user interaction. 
Since text alone is often insufficient to precisely express complex or fine-grained user intent, \citet{wang2026virtuevisualinteractivetextimageuniversal} enable visual prompts as an additional human-machine interaction modality to embedders, yet their approach still operates within the image-text paradigm.
We argue that humans should be able to interact with embedding models not only through textual and visual guidance, but also via video and audio modalities as interaction modalities, which we term \textit{omni-interactions}.
For instance, visual prompts may fall short in video scenarios involving both on-screen and off-screen sources or ambiguous audio-visual correspondences, where selecting a spatial region alone cannot reliably capture the intended semantics.
In addition, in complex auditory scenes with overlapping or intermittent sound sources (e.g., retrieving segments that contain periodic bird chirps embedded in background noise, rather than all clips matching the coarse label "bird"), textual queries are often insufficiently precise to capture retrieval intent, while visual cues may be absent or weakly correlated, necessitating direct audio-based interaction.
Enabling omni-interactions in embedding models not only facilitates more accurate and user-aligned representations, but also advances toward truly universal embedders for user-conditioned interactions, with applications beyond retrieval, including context engineering \cite{contextengineering}, agentic systems \cite{openclaw,DBLP:journals/corr/abs-2504-07079,DBLP:journals/corr/abs-2601-01743}, LLM memory augmentation \cite{DBLP:journals/corr/abs-2509-25140,du2025rethinkingmemoryllmbased}, and personalization \cite{liang2026learningpersonalizedagentshuman}.
These challenges give rise to two central yet underexplored questions: \textit{1) How can we enable omni-interactions in omni-modal embedding models? 2) Given such capabilities, how can we systematically evaluate individual or joint interactions within multimodal embedding benchmarks?}

\begin{figure}[t]
    \centering
    \includegraphics[width=\textwidth]{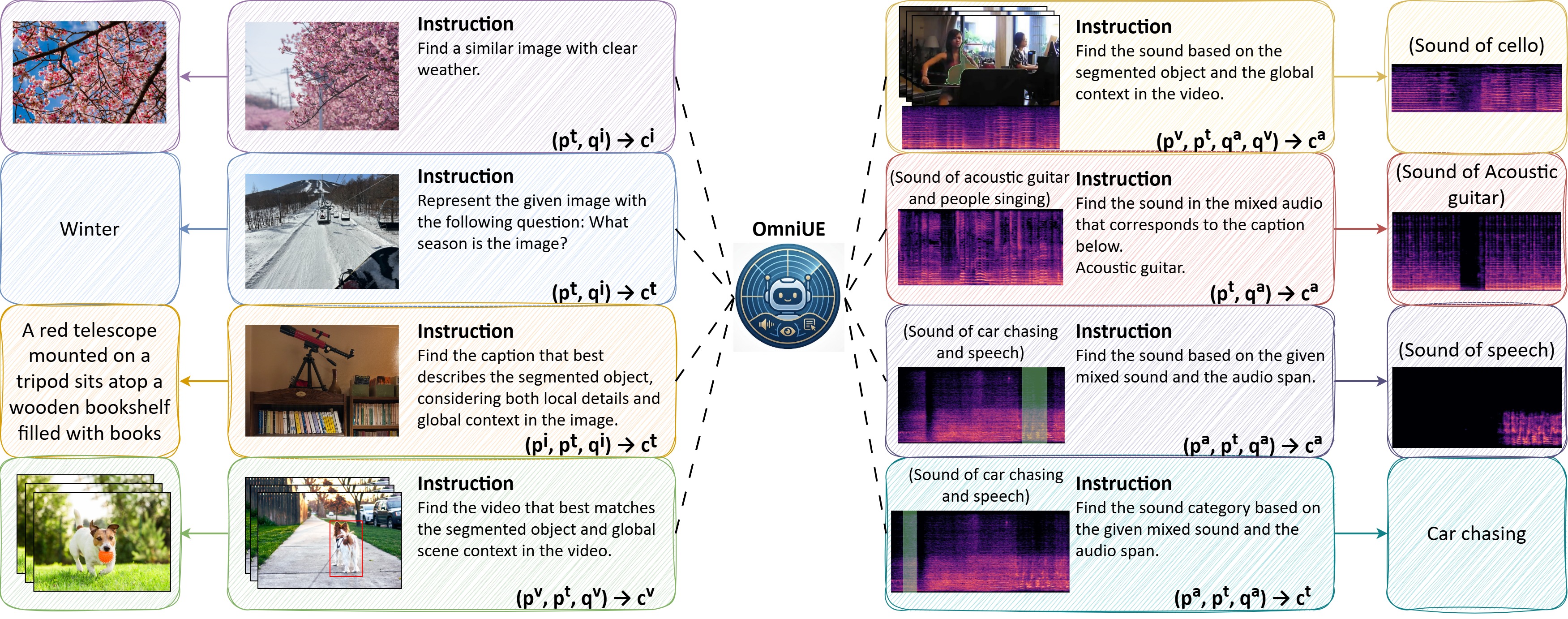}
    \vspace{-15pt}
    \caption{
    Applicability of OmniUE, which takes omni-interactive prompts ($p$) and queries ($q$) to retrieve any-modality targets ($c$), including text, image, video, and audio.}
    \label{fig:motivation-example}
\end{figure}

In this paper, we propose OmniUE, a novel omni-interactive universal embedder that supports optional omni-modal interactions and arbitrary combinations of text, silent video, and audio inputs to produce a unified representation.
Specifically, OmniUE is built upon two complementary axes: i) a query stream, which encodes multimodal inputs (visual, textual, and audio) into holistic contextual representations, and (ii) an interactive prompt stream, which captures user-specified signals as entity- or intent-level guidance.
To learn multi-level context across complex multi-modalities, we introduce additional learnable tokens appended in the sequence and aggregate them from intermediate LLM layers into a final single embedding.
These axes enable flexible and fine-grained conditioning of embeddings through heterogeneous interaction signals, while maintaining a unified representation space.
To the best of our knowledge, OmniUE is the first framework to generate user-conditioned unified embeddings across arbitrary modality combinations (as shown in Fig. \ref{fig:motivation-example}), generalizing beyond prior approaches that rely primarily on textual instructions\footnote{We treat text as an interaction prompt ($p^t$) unless otherwise specified, due to its ambiguous role.}.
Importantly, such textual interaction can be viewed as a special case within our broader omni-interaction paradigm, highlighting the generality and extensibility of our formulation.

Since existing embedding benchmarks primarily evaluate textual interactions on video-text and audio-text pairs, and visual interactions on image-text pairs, they fail to capture the full spectrum of omni-modal inputs and to support both unimodal and multimodal interactions.
To evaluate audio interactions as well as video-audio alignment, we introduce OmniCHOIR, an omni-interactive text-video-audio-to-audio (TVA2A) retrieval benchmark that systematically evaluates textual, visual, and audio interactions.
In OmniCHOIR, given a video with mixed audio and a query (which may consist of a single modality or a combination of modalities), the task is to retrieve the sound that matches the intended target from the mixture, conditioned on the corresponding background context.
To further increase compositional and reasoning difficulty, we construct multiple hard negative distractors with different background sounds, the same background but containing sounds from the same target category, and both different sound categories and backgrounds.
This design establishes a unique property of OmniCHOIR: it evaluates embedding models not only across diverse interaction modalities, but also under fine-grained compositional and reasoning challenges, providing a more rigorous and realistic testbed for omni-interactive embeddings.

In summary, our contributions are three-fold:
\begin{tightitem}
    \item \textbf{Method Novelty:} We propose OmniUE, the first omni-interactive omni-modal universal embedder that accepts text, visual, and audio prompts as interactions jointly with arbitrary combinations of text, visual, and audio inputs via segmenters and multi-layer context aggregation to generate a unified embedding. The visual segmenter allows users to provide different types of visual prompts, and the audio segmenter enables users to provide temporal audio spans. 
    \item \textbf{Benchmark Novelty:} Since there is no omni-interactive embedding benchmark, we introduce OmniCHOIR, composed of omni-interactive TVA2A retrieval that aims to assess embedders' reasoning, compositional, and unimodal or multimodal user-conditioned capabilities. OmniCHOIR enables advancements on omni-modal interaction scenarios that remain unexplored.
    \item \textbf{Experiment Novelty:} OmniUE significantly and consistently outperforms state-of-the-art embedding baselines across multiple benchmarks, including 18 text-interactive video tasks (MMEB-v2-video), 30 text-interactive audio tasks (MAEB), 5 visual-interactive text-image-to-text tasks (SCaR), and the omni-interactive OmniCHOIR benchmark, achieving average improvements of 10.4\%, 1.0\%, 83.1\%, and 21.9\%, respectively.
\end{tightitem}
\section{Related Works}
\label{sec:related-work}

\noindent\textbf{LLM-Based Multimodal Embedders.}
Most existing works focus on advancing MLLM-based embedders \cite{DBLP:journals/corr/abs-2407-12580,DBLP:conf/iclr/JiangMYYZC25,DBLP:conf/cvpr/ZhangZXLDLXZLZ25,DBLP:conf/cvpr/LiuZCJ0YWX25,DBLP:journals/corr/abs-2511-19278}, which demonstrate effective performance and instruction-following capabilities for handling diverse user queries in embedding tasks, outperforming traditional dual-encoder paradigms, e.g., \cite{DBLP:conf/icml/RadfordKHRGASAM21,DBLP:conf/emnlp/XuG0OAMZF21,DBLP:conf/icml/0001LXH22,DBLP:conf/iccv/ZhaiM0B23,DBLP:conf/cvpr/ChertiBWWIGSSJ23,DBLP:conf/icml/0033LHLQC0C24,DBLP:conf/eccv/WeiCCHZFRC24,DBLP:journals/corr/abs-2504-17432}.
In parallel, audio-language representation learning has been advanced by works such as \cite{DBLP:conf/icassp/ElizaldeDIW23}, which introduce CLAP to align audio and text in a shared embedding space for classification and retrieval.
Subsequent efforts further improve performance by scaling training data and enhancing caption quality for fine-grained audio-text alignment \cite{DBLP:conf/icassp/WuCZHBD23,DBLP:journals/taslp/MeiMLKKZPZW24,bai2024audiosetcapsnipsws,DBLP:conf/icassp/YuanJZCC000KP025}.
While these methods advance the realm of multimodal representation learning, the primary hurdles lie in their focus on text paired with a single additional modality and in their reliance on text as the sole interface for user-conditioned interaction.
Recently, omni-modal embedders have been proposed to unify visual (image and video), audio, and text within a shared embedding space, either through binding modalities \cite{DBLP:conf/cvpr/GirdharELSAJM23,DBLP:conf/iclr/ZhuLNYCWPJZLZ0024} or omni-modal LLMs \cite{xiao2025scaling,DBLP:journals/corr/abs-2510-03458,DBLP:journals/corr/abs-2601-03666,DBLP:conf/sigir/MaGZZCL25,tang2026wave,DBLP:journals/corr/abs-2603-02098}; nonetheless, these approaches continue to rely on text as the sole interaction modality.
To date, \cite{wang2026virtuevisualinteractivetextimageuniversal} is the only work that extends interaction beyond text by introducing visual prompts; yet, it is limited to the image-text setting and does not generalize to video or audio, nor does it support audio-based interactions.
To address these limitations, our proposed OmniUE is able to accommodate not only omni-modal inputs, but also any individual modality (or their combinations) as interaction media by introducing additional video and audio segmenters and context aggregation that aims to incorporate complex information.

\noindent\textbf{Interactive Embedding Benchmarks.}
With the rapid advancement of MLLM-based embedders for diverse embedding tasks, evaluation benchmarks have shifted from traditional unimodal (e.g., BEIR \citep{DBLP:conf/nips/Thakur0RSG21}, MTEB \citep{DBLP:conf/eacl/MuennighoffTMR23}) and bimodal retrieval settings to large-scale, instruction-driven multimodal benchmarks that assess instruction-following capabilities across diverse scenarios.
\citet{DBLP:journals/corr/abs-2507-04590} introduce MMEB-v2-video, comprising 18 text-video embedding tasks spanning video QA, classification, retrieval, and moment retrieval, by reformulating both retrieval and generative tasks into embedding tasks with different textual instructions.
MAEB \cite{DBLP:journals/corr/abs-2602-16008} provides a large-scale text-audio embedding benchmark covering 30 tasks across speech, music, environmental sounds, and cross-modal reasoning in over 100 languages.
As existing benchmarks primarily rely on textual instructions, SCaR \cite{wang2026virtuevisualinteractivetextimageuniversal} introduces a large-scale visual-interactive text-image-to-text (TI2T) retrieval benchmark, composed of five visual grounding datasets for region-level caption retrieval within global scene context.
While we leverage MMEB-v2-video, MAEB, and SCaR to extensively evaluate textual and visual interactive embedding capabilities, these benchmarks do not assess scenarios involving video-audio alignment or audio-based interactions, where audio spans themselves serve as prompts.
To fill this gap, we introduce OmniCHOIR, an omni-interactive text-video-audio-to-audio (TVA2A) retrieval benchmark that challenges models with unimodal and multimodal interactions.

\section{OmniUE}

\begin{figure}[t]
    \centering
    \includegraphics[width=\textwidth]{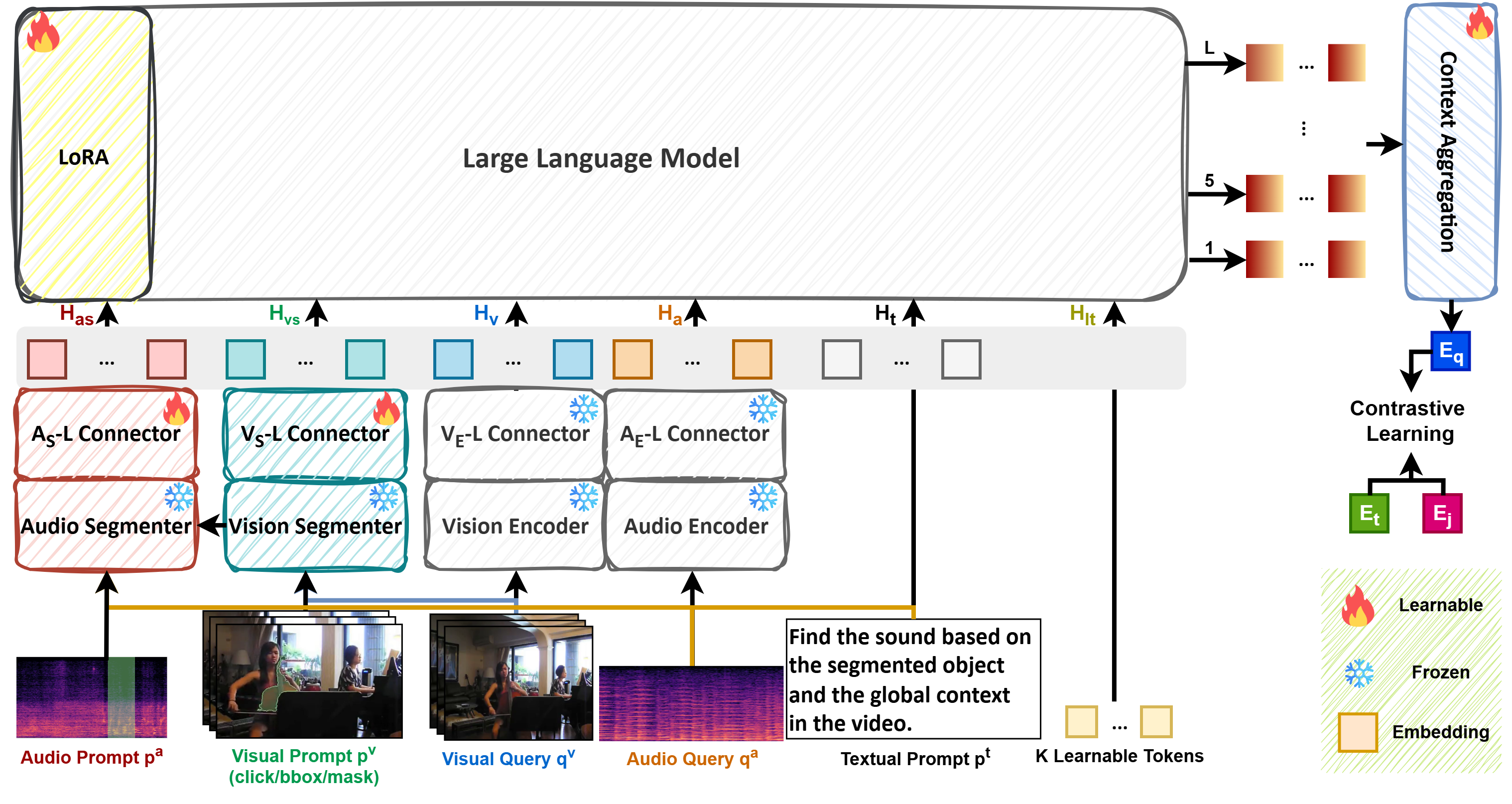}
    \vspace{-15pt}
    \caption{
    OmniUE overview. Compared to existing omni-LLM-based embedders (shown in gray; e.g., \cite{xiao2025scaling}), OmniUE introduces three key components: 1) audio and visual segmenters that enable emerging interaction capabilities; 2) additional learnable tokens for modeling complex and compositional multimodal inputs; and 3) a context aggregation module that selectively weighs the token-layer importance of contextual tokens to produce the final embedding for contrastive learning.}
    \label{fig:model-framework}
\end{figure}

\subsection{Overview}
Fig.~\ref{fig:model-framework} presents an overview of our proposed OmniUE, which supports text-only, video-only ($q^v$), audio-only ($q^a$), and joint multimodal inputs, along with visual-, audio-, and textual interaction prompts ($p^v, p^a, p^t$).
Visual and audio inputs, including both queries and interaction prompts, are processed by their respective segmenters/encoders and connectors to produce user-conditioned holistic hidden representations.
In addition, $K$ learnable tokens are appended to the input sequence and jointly processed by the LLM to capture compositional and complex multimodal information, which has been found to be useful for the image-text paradigm \citep{DBLP:journals/corr/abs-2511-19278}.
The hidden states of these $K$ tokens are extracted at a fixed interval of $l$ layers and aggregated via a context aggregation module to produce the final embedding $E$.
The query ($E_q$), target ($E_t$), and negative ($E_j$) embeddings are then jointly optimized using the InfoNCE loss \citep{DBLP:journals/corr/abs-1807-03748}.
We adopt GradCache \citep{DBLP:conf/rep4nlp/GaoZHC21} to enable larger batch sizes and improve generalization over in-batch negatives, following prior work \citep{wang2026virtuevisualinteractivetextimageuniversal}.

\noindent\textbf{Converting modality inputs into embeddings.}
The omni-LLM consists of vision and audio encoders and their corresponding connectors.
Given a multimodal input (either a query or a target) that comprises any combination of text, visual, and audio modalities, the resulting tokenized $d$-dimensional embeddings for each modality are denoted as $H_v$, $H_a$, and $H_t$, respectively.

\noindent\textbf{Appending additional multi-token inputs.}
A common practice to obtain a single representation $E$ is to use the final hidden state of the EOS token from the last layer, as is widely adopted in MLLM embedders \citep{DBLP:journals/corr/abs-2507-04590,wang2026virtuevisualinteractivetextimageuniversal,xiao2025scaling}.
However, naively relying on the last-layer representation of a single token may degrade the inherent capabilities of MLLMs \citep{DBLP:conf/cvpr/OualiBXZMMT25}, and a single token is often insufficient to encode complex omni-modal information.
To extend the multi-token design to the compositionally denser omni-modal realm, we augment each input sequence with $K=4$ learnable tokens $H_{lt} \in \mathbb{R}^{K \times d}$.
These tokens are randomly initialized and trained to capture embedding-specific information beyond the original MLLM vocabulary.
Nevertheless, this design remains restricted to using hidden states from the final layer; we address this limitation in Sec.~\ref{sec:aggregation}.

\subsection{Enabling Emerging Visual- and Audio-Interactive Capabilities}
While $H_v$ and $H_a$ encode holistic representations of visual and audio inputs, they are insufficient for incorporating user-specified interaction prompts and capturing fine-grained information (e.g., entity-level cues for vision and temporal span-level cues for audio).
To address this limitation, we introduce two additional segmenter streams for vision and audio, which support points, bounding boxes, masks, text, and temporal spans as interaction inputs, producing the corresponding interaction embeddings $H_{vs}$ and $H_{as}$.

\noindent\textbf{Encoding interactive prompts into interaction embeddings.}
We adopt pretrained SAM-3 \citep{carion2025sam3segmentconcepts} as the vision segmenter since it effectively segments entities based on user-specified regions of interest.
In parallel, SAM-Audio \citep{shi2025samaudio} serves as the audio segmenter, enabling sound separation from mixed audio conditioned on optional video, text, or audio temporal spans.
A key advantage is that SAM-3 can be viewed as a submodule of SAM-Audio, allowing them to be integrated into a unified interaction framework.
Specifically, we extract the visual feature map $F_{vs} \in \mathbb{R}^{|F^H_{vs}| \times |F^W_{vs}| \times d_{vs}}$ from before the mask prediction head in SAM-3, which integrates video inputs from the image encoder and visual prompts from the detector.
Similarly, the audio feature map $F_{as} \in \mathbb{R}^{|F_{as}| \times d_{as}}$ is obtained before the decoder (i.e., after the Diffusion Transformer \citep{DBLP:conf/iccv/PeeblesX23} and the iterative sampling procedure) in SAM-Audio.
We use feature maps prior to decoding since they already encode interaction-aware information from both the inputs and interaction prompts, eliminating the need for additional decoding back into embedding space.
For non-visual interactive scenarios, visual prompts are omitted, directly yielding the corresponding feature maps.
For non-audio interactive scenarios, prompts are set to their default forms, namely zero vectors for videos, \texttt{<null>} tokens for spans, and empty strings for text, following the design of SAM-Audio.
Therefore, segmenter-derived embeddings are employed even in purely text-interactive settings, enabling the model to capture entity-level cues while preserving the global context from the vision encoder within the omni-LLM.

\noindent\textbf{Connecting interaction embeddings to the unified LLM space.}
Directly flattening visual and audio feature maps leads to significantly long sequences when aligning with the LLM (e.g., $|F^H_{vs}| \times |F^W_{vs}|=288\times288$ in SAM-3).
To this end, we introduce segmenter connectors by applying a $\text{Conv2D}$ layer to reduce the spatial resolution of visual feature maps and a $\text{Conv1D}$ layer to project audio feature maps into the LLM-compatible sequence space.
These representations are then passed through $h$-head self-attention connectors ($\text{SA}$) and $\text{MLP}$ to produce the final interaction embeddings:
\begin{equation}
\small
    H_{vs} = \text{MLP}_{vs}(\text{SA}_{vs}(\text{Conv2D}(F_{vs}))) \in \mathbb{R}^{|F'_{vs}| \times d},
    H_{as} = \text{MLP}_{as}(\text{SA}_{as}(\text{Conv1D}(F_{as}))) \in \mathbb{R}^{|F'_{as}| \times d}, \
\end{equation}
where $|F'_{vs}|$ and $|F'_{as}|$ are the compressed lengths of $H_{vs}$ and $H_{as}$, respectively.

\subsection{Aggregating Contextual and Intermediate Tokens to the Final Embedding}
\label{sec:aggregation}
A key challenge in omni-modal embedding is how to effectively aggregate heterogeneous signals, especially for multimodal inputs and user interactions.
Relying solely on the final-layer representation, as in prior work (e.g., \cite{xiao2025scaling,DBLP:conf/cvpr/OualiBXZMMT25}), overlooks multi-level or modality-specific information captured at intermediate depths.
In contrast, MLP-based layer aggregation methods (e.g., \cite{tang2026wave}) introduce substantial parameter overhead due to the high dimensionality of LLM hidden states, with a complexity of $O(L \cdot d^2)$, where $L$ is the number of layers.
For instance, fine-tuning Qwen2.5-omni-3B ($d$ is 2048 and $L$ is 36) \citep{Qwen2.5-Omni} with LoRA requires around 45M trainable parameters, whereas the MLP used for late fusion alone introduces 155M additional parameters, potentially leading to underfitting (see Appendix \ref{app:ablation-study}).

To address this, we first construct a unified sequence by concatenating holistic context, user-conditioned representations, and learnable tokens as $H = [H_{as}, H_{vs}, H_v, H_a, H_t, H_{lt}]$, which is then fed into the LLM.
To capture multi-level representations, we extract the hidden states of the learnable tokens $H_{lt}$ every $l$ layers (with $l=4$) starting from the first (i.e., layers 1, 5, 9, $\dots$), and include the final layer if not already selected.
Then, we introduce a context aggregation module to fuse these representations.
Let $Z \in \mathbb{R}^{M \times K \times d}$ denote the collected $K$ token features across the $M$ selected layers.
Instead of treating all layers equally, we introduce a set of normalized layer-token importance weights $W \in \mathbb{R}^{M \times K}$, allowing the model to learn a global preference over representation levels and tokens, inspired by \cite{DBLP:conf/naacl/PetersNIGCLZ18}.
The final embedding $E \in \mathbb{R}^d$ is:
\begin{equation}
E = \phi(\tilde{Z}), \tilde{Z} = \sum_{m=1}^M \tilde{W}_m \odot Z_m \in \mathbb{R}^{K \times d}, \tilde{W}=\text{softmax}(W), \tilde{W}_m \in R^K,
\end{equation}
where $\odot$ denotes element-wise multiplication with broadcasting along the feature dimension, $\phi$ is mean pooling over the $K$ tokens, and the softmax operation is applied across both the layer and token dimensions.
In this manner, the model selectively integrates contextual and optionally interaction-aware signals from both local and global representations retrieved from earlier and later LLM layers, while maintaining computational efficiency.

\section{OmniCHOIR: Omni-Interactive TVA2A Retrieval Benchmark}

\begin{figure}[t]
    \centering
    \includegraphics[width=\textwidth]{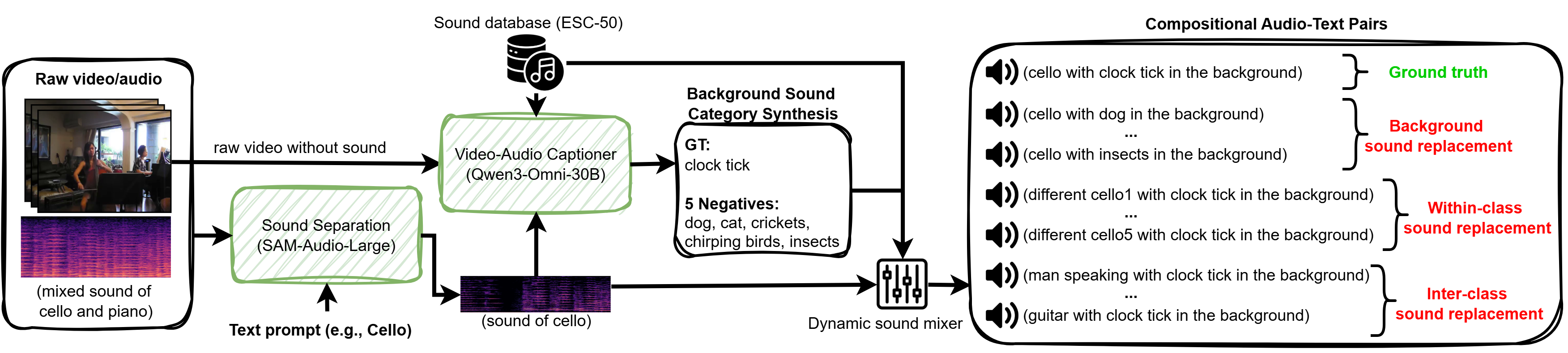}
    \vspace{-10pt}
    \caption{
    The data collection pipeline for OmniCHOIR. We use SAM-Audio-Large for sound separation conditioned on text prompts, and Qwen3-Omni-30B as a captioner to synthesize ground-truth and negative background sound categories based on the visual information from the silent video. A dynamic sound mixer then injects background sounds into the target audio track with randomized temporal intervals and volume levels (-15 dB to -5 dB).}
    \label{fig:omnichoir}
\end{figure}

\subsection{Overview}
As discussed in Sec. \ref{sec:related-work}, existing benchmarks primarily focus on textual-interactive embedding evaluation, while only \cite{wang2026virtuevisualinteractivetextimageuniversal} introduces a visual-interactive benchmark, which is limited to the text-image paradigm.
To complement existing benchmarks and move toward a more comprehensive evaluation, we introduce OmniCHOIR, an omni-interactive text-video-audio-to-audio retrieval benchmark.
We focus on TVA2A retrieval because audio mixtures inherently involve the superposition of multiple sources, leading to ambiguities that are qualitatively more challenging than visual occlusion, and thus require precise multimodal interaction (e.g., temporal spans or cross-modal cues) to resolve.
Moreover, compared to visual region-based interaction, audio-centric interaction remains unexplored in existing benchmarks.
OmniCHOIR challenges models not only in reasoning and compositionality, but also in handling diverse combinations of multimodal interaction prompts.
Each instance in OmniCHOIR comprises a raw video, its synchronized mixed audio, visual segmentation masks for the target subject, audio temporal spans corresponding to the target sound, and textual descriptions of the desired sound.

\noindent\textbf{Problem Definition.}
Given a raw video $V$, the corresponding mixed audio $A$, and an interaction prompt $P$, where $P$ can be any combination of $p^a$, $p^v$, and $p^t$, the goal is to retrieve the most relevant audio that matches both the target sound and its background context from a set of 16 audio candidates $C$: $c_{y} = \operatorname*{argmax}_{c_i \in C} \cos\!\left(\theta(V, A, P),\, \theta(c_i)\right), C=\{c_1,\dots,c_{16}\},$
where $\theta$ denotes the embedding model.
In practice, textual instructions are always included (i.e., \textit{"Find the sound from the mixed audio and the video: \{text condition\}"} for $p^t$, \textit{"Sound occurring between \{start\}s and \{end\}s"} for $p^a$, and \textit{"Identify the sound from the highlighted region"} for $p^v$), and are omitted here for simplicity.

\subsection{Collection Pipeline}
\label{sec:collection-pipeline}
Fig.~\ref{fig:omnichoir} illustrates the pipeline for constructing OmniCHOIR.
We build upon SAM-Audio-Bench\footnote{It is not included when training SAM-Audio.} \cite{shi2025samaudio} as the base dataset, which contains 819 10-second samples with 24 fps video $V$, audio $A$, and multimodal interaction prompts $P$.
SAM-Audio-Bench is a real-world multimodal separation benchmark with diverse taxonomic coverage, including speech cleaning, speaker separation, music removal, instrument stems (37 classes), and general environmental sounds.
Each instance provides human-annotated visual masklets (for on-screen sounding objects), temporal spans, and textual descriptions as interaction signals $P$.
The video and audio data are sourced from six public datasets: AudioSet \citep{DBLP:conf/icassp/GemmekeEFJLMPR17}, VGGSound \citep{DBLP:conf/icassp/ChenXVZ20}, MUSIC \citep{DBLP:conf/eccv/ZhaoGRVMT18}, MUSIC-AVQA \citep{DBLP:conf/cvpr/LiWTXW022}, AVSpeech \citep{DBLP:journals/tog/EphratMLDWHFR18}, and CondensedMovies \citep{DBLP:conf/accv/BainNBZ20}.
Importantly, all samples in SAM-Audio-Bench are held out, ensuring that OmniUE was not exposed to these data during training and thereby supporting evaluation of zero-shot generalization.

Since SAM-Audio-Bench does not provide clean target sounds, we apply SAM-Audio-Large \cite{shi2025samaudio} to perform sound separation, followed by manual validation.
After filtering out invalid samples and missing videos, we retain 479 instances.
Next, we employ Qwen3-Omni-30B \cite{Qwen3-Omni} as a captioner, conditioned on the silent video and the target sound, to generate one ground-truth and five negative background-sound categories based on the ESC-50 sound taxonomy \cite{piczak2015dataset}.
The prompt instructs the model to select plausible background sounds for the scene that are not strongly supported by the visual context and are semantically distinct from a ground-truth category.
We also encourage Qwen3-Omni to produce diverse and creative negatives, ensuring coverage across different categories, while reducing the risk of overfitting to narrow patterns.

We then introduce a dynamic sound mixer, which combines the target sound with the ground-truth background sound to form the reference audio.
The background sounds are randomly sampled from the ESC-50 dataset based on the synthesized categories and injected at randomized temporal intervals.
To ensure acoustic diversity and realism, we vary the relative volume levels of the background sounds within a range of -15 dB to -5 dB relative to the target signal.
To construct challenging negative candidates, we design three replacement strategies, each producing five candidates to balance difficulty and diversity:
\textbf{1) Background sound replacement:} replace only the background sound using the five generated negatives.
\textbf{2) Within-class sound replacement:} replace the target sound with another separated sound from the same category within SAM-Audio-Bench.
\textbf{3) Inter-class sound replacement:} replace the target sound with another target sound from a different category.
All corresponding text descriptions, which serve as OmniCHOIR's text conditions, are generated using the template \textit{"\{targeted sound\} with \{background sound\} in the background."}, forming the final instances of OmniCHOIR.
Additional prompt templates and examples are provided in Appendix~\ref{app:omnichoir-details}.
\begin{table}[t]
    \centering
    \caption{Results on 18 MMEB tasks, with scores averaged per meta-task. \hlpink{Improvements} are computed against the best-performing 2B/3B and 7B baselines. We highlight the \textbf{best} and \underline{second-best} in each column. All results are from the leaderboard and \cite{DBLP:journals/corr/abs-2601-03666}, except for LCO-Emb, which we reproduced.}
    \setlength{\tabcolsep}{20pt} 
    \scalebox{0.7}{
        \begin{tabular}{lccccc}
    \toprule
      & CLS & QA & RET & MRET & \textbf{Overall} \\
     \midrule
     \textbf{Model $\downarrow$ \; $\mid$ \; \#Datasets $\rightarrow$} & 5 & 5 & 5 & 3 & 18 \\
    \midrule
    \multicolumn{6}{c}{\textbf{Text + Image + Video}} \\
    \midrule
    VLM2Vec-v2 (Qwen2-VL-2B) \cite{DBLP:journals/corr/abs-2507-04590} & 39.3 & 34.3 & 28.8 & 38.5 & 34.9 \\
    \rowcolor{gray!10}
    UME-R1 (Qwen2-VL-2B) \cite{DBLP:journals/corr/abs-2511-00405} & 44.3 & 51.2 & 32.9 & 39.7 & 42.2 \\
    \hdashline
    UniME-V2 (LLaVA-OneVision-7B) \cite{DBLP:journals/corr/abs-2510-13515} & 37.2 & 50.6 & 28.9 & 39.6 & 39.0 \\
    \rowcolor{gray!10}
    UME-R1 (Qwen2-VL-7B) \cite{DBLP:journals/corr/abs-2511-00405} & 48.6 & \underline{60.7} & \underline{38.2} & 39.3 & 47.5 \\
    CAFe (LLaVa-OneVision-7B) \cite{DBLP:conf/iccv/YuZYKWHLZFY25} & 35.8 & 58.7 & 34.4 & 39.5 & 42.4 \\

    \midrule
    
    \multicolumn{6}{c}{\textbf{Text + Image + Video + Audio (Omni-modal)}} \\
    \midrule
    Omni-Embed-Nemotron (Qwen2.5-omni-3B) \cite{DBLP:journals/corr/abs-2510-03458} & 40.5 & 44.3 & 32.7 & 25.6 & 36.9 \\
    \rowcolor{gray!10}
    LCO-Emb (Qwen2.5-omni-3B) \cite{xiao2025scaling} & 42.9 & 56.7 & 30.0 & 43.8 & 43.3 \\
    e5-omni (Qwen2.5-omni-3B) \cite{DBLP:journals/corr/abs-2601-03666} & 40.2 & 48.5 & 33.2 & 40.7 & 40.6 \\
    \hdashline
    \rowcolor{gray!10}
    LCO-Emb (Qwen2.5-omni-7B) \cite{xiao2025scaling} & 39.3 & 57.6 & 24.8 & 26.5 & 38.2 \\
    e5-omni (Qwen2.5-omni-7B) \cite{DBLP:journals/corr/abs-2601-03666} & 46.6 & 52.9 & 36.7 & 34.2 & 43.5 \\
    \rowcolor{gray!10}
    WAVE (Qwen2.5-omni-7B)  \cite{tang2026wave} & 50.7 & 45.9 & 34.7 & 39.5 & 43.1 \\
    
    \midrule
    \rowcolor{sonyblue!15}
    \multicolumn{6}{c}{\textbf{Ours (Omni-modal)}} \\
    \midrule
    OmniUE-3B disabling audio \& vision segmenters & 47.8 & 52.6 & 34.5 & \underline{45.1} & 45.0 \\
    \rowcolor{gray!10}
    OmniUE-7B disabling audio \& vision segmenters & \underline{54.7} & 59.9 & 37.6 & 35.5 & 48.2 \\
    \hdashline
    OmniUE-3B & 51.1 & 57.8 & 36.9 & \textbf{47.7} & \underline{48.4} \\
    \rowcolor{gray!10}
    OmniUE-7B & \textbf{57.8} & \textbf{62.9} & \textbf{40.9} & 41.3 & \textbf{51.8} \\

    \midrule

    \rowcolor{pink!30!white}
    Improvements (3B) & +6.8 & +1.1 & +3.7 & +3.9 & +5.1 \\
    \rowcolor{pink!30!white}
    Improvements (7B) & +9.2 & +2.2 & +2.7 & +1.7 & +4.3 \\
    \bottomrule
\end{tabular}

    }
    \label{tab:mmeb-results}
\end{table}

\section{Experiments}
\subsection{Experimental Setup}
\label{sec:setup}
\noindent\textbf{Benchmarks.}
To comprehensively evaluate OmniUE's capabilities, we conduct experiments on textual, visual, and omni-interactive benchmarks.
Specifically, \textbf{MMEB-v2-video} \cite{DBLP:journals/corr/abs-2507-04590} assesses 18 textual-interactive video tasks, including classification (CLR), question answering (QA), retrieval (RET), and moment retrieval (MRET).
\textbf{MAEB} \cite{DBLP:journals/corr/abs-2602-16008} examines 30 textual-interactive audio tasks, including classification (Clf), multi-label classification (M.Clf), pairwise classification (PC), reranking (Rrnk), clustering (Clust), audio retrieval (A.Rtrvl), cross-modal retrieval (X.Rtrvl), and zero-shot classification (Zero Clf.).
\textbf{SCaR} \cite{wang2026virtuevisualinteractivetextimageuniversal} evaluates 5 visual-interactive text-image-to-text (TI2T) retrieval tasks across different domains.
In addition, our proposed \textbf{OmniCHOIR} assesses omni-interactive TVA2A retrieval tasks that consist of text, visual regions, and audio temporal spans as interaction media.
We use the same metrics as the official benchmarks, and Recall@1 is used for OmniCHOIR.

\noindent\textbf{Implementation Details.}
We use 3B and 7B checkpoints of Qwen2.5-Omni \cite{Qwen2.5-Omni} as the omni-LLM backbone for OmniUE-3B and OmniUE-7B, respectively, both of which utilize sam-audio-base \cite{shi2025samaudio} equipped with SAM-3 \cite{carion2025sam3segmentconcepts} as the audio and vision segmenters.
LoRA \cite{DBLP:conf/iclr/HuSWALWWC22} is applied for the LLM with the rank set to 32 and alpha set to 64, and the audio prompt and visual prompt connectors are trained from scratch to align the segmenter space to the omni-LLM space.
OmniUE is trained on around 3.5M multimodal samples from public datasets, with a batch size of 1024.
To evaluate whether the interaction streams remain beneficial even when no explicit visual or audio interaction prompts are available, disabling the visual \& audio segmenters denotes the corresponding interaction streams being disabled only during inference (i.e., no inputs to visual and audio interaction streams), while the model is still trained with them enabled with default inputs as described in Sec. 3.2.
To evaluate the emerging visual- and audio-interactive capabilities, we do not include the corresponding interactions in the training datasets, i.e., only textual-interactive datasets are used for training OmniUE.
Detailed datasets and configurations are provided in Appendix \ref{app:implementation-details}, with ablation and parameter studies, and inference time and memory analyses in Appendices \ref{app:ablation-study} and \ref{app:inference-analysis}, respectively.

\subsection{Textual-Interactive Performance}
\noindent\textbf{Video benchmarks (MMEB-v2-video).}
Tab.~\ref{tab:mmeb-results} summarizes the overall performance on MMEB-v2-video.
OmniUE consistently outperforms all baselines, including text+image+video and omni-modal models, across all four meta-task scenarios.
Quantitatively, OmniUE-3B surpasses existing 2B/3B models by 5.1 points on average, while OmniUE-7B improves over the 7B baselines by 4.3 points.
To analyze the source of these gains, we compare OmniUE with variants that remove the audio and visual segmenter streams.
The performance drop in these variants indicates that incorporating fine-grained, entity-level information is critical for textual-interactive embedding.
Notably, the segmenter streams remain beneficial even when no explicit interaction prompts are provided, suggesting that they complement global multimodal representations with localized details via segmenter connectors.

\begin{table}[t]
    \centering
    \caption{Results on 30 MAEB tasks. Scores are averaged per meta-task. \hlpink{Improvements} are computed relative to the best-performing 3B and 7B baselines. We highlight the \textbf{best} and \underline{second-best} results in each column. Baseline results are from the MAEB leaderboard.}
    \setlength{\tabcolsep}{8.5pt} 
    \scalebox{0.65}{
        \begin{tabular}{lccccccccc}
    \toprule
      & Clf & M.Clf & PC & Rrnk & Clust & A. Rtrvl & X. Rtrvl & Zero Clf. & \textbf{Overall} \\
     \midrule
     \textbf{Model $\downarrow$ \; $\mid$ \; \#Datasets $\rightarrow$} & 10 & 2 & 3 & 1 & 3 & 1 & 8 & 2 & 30 \\
    \midrule
    \multicolumn{10}{c}{\textbf{Two-Tower-Based Embedders}} \\
    \midrule
    microsoft/msclap-2023 \cite{DBLP:conf/icassp/ElizaldeDIW23} & 45.0 & 5.8 & 53.6 & 75.4 & \textbf{15.2} & 87.3 & 9.4 & 12.6 & 31.1 \\
    \rowcolor{gray!10}
    laion/larger\_clap\_general \cite{DBLP:conf/icassp/WuCZHBD23} & 51.7 & 2.3 & 51.9 & 66.8 & 6.6 & \underline{93.2} & 9.8 & 14.9 & 32.2 \\

    \midrule
    
    \multicolumn{10}{c}{\textbf{LLM-Based Embedders}} \\
    \midrule
    LCO-Emb (Qwen2.5-Omni-3B) \cite{xiao2025scaling} & 56.4 & 41.6 & 66.7 & 75.4 & 1.3 & 67.7 & \textbf{50.3} & 62.2 & 50.7 \\
    \rowcolor{gray!10}
    Qwen2-Audio-7B \cite{DBLP:journals/corr/abs-2407-10759} & \textbf{62.7} & 10.7 & 56.9 & 80.8 & \underline{12.7} & 33.9 & 1.6 & 12.4 & 33.7 \\
    LCO-Emb (Qwen2.5-Omni-7B) \cite{xiao2025scaling} & 58.0 & \underline{45.7} & 67.3 & 78.7 & 1.7 & 78.2 & \textbf{50.3} & 64.5 & \underline{52.2} \\
    
    \midrule
    \rowcolor{sonyblue!15}
    \multicolumn{10}{c}{\textbf{Ours (Omni-modal)}} \\
    \midrule
    OmniUE-3B disabling audio \& vision segmenters & 53.6 & 20.5 & 67.3 & 87.0 & 1.3 & 80.4 & 44.7 & 56.5 & 45.1 \\
    \rowcolor{gray!10}
    OmniUE-7B disabling audio \& vision segmenters & 59.0 & 23.5 & \underline{68.5} & \underline{88.3} & 1.8 & 92.4 & 24.5 & \underline{66.6} & 45.5 \\
    \hdashline
    OmniUE-3B & 57.3 & 35.3 & 67.5 & \textbf{88.4} & 1.7 & 81.1 & 46.5 & 63.5 & 50.5\\
    \rowcolor{gray!10}
    OmniUE-7B & \underline{59.1} & \textbf{49.9} & \textbf{68.8} & 88.2 & 2.7 & \textbf{93.3} & \underline{47.7} & \textbf{67.1} & \textbf{53.4} \\

    \midrule

    \rowcolor{pink!30!white}
    Improvements (3B) & +0.9 & - & +0.8 & +13.0 & - & +13.4 & - & +1.3 & - \\
    \rowcolor{pink!30!white}
    Improvements (7B) & - & +4.2 & +1.5 & +7.4 & - & +15.1 & - & +2.6 & +1.2 \\
    \bottomrule
\end{tabular}

    }
    \label{tab:maeb-results}
\end{table}

\noindent\textbf{Audio benchmarks (MAEB).}
Beyond video evaluation, Tab.~\ref{tab:maeb-results} reports results on 30 textual-interactive audio tasks from MAEB.
OmniUE consistently outperforms both two-tower and LLM-based embedders, achieving substantial gains across all meta-tasks.
On average, OmniUE-3B achieves on par performance with LCO-Emb-3B.
Compared to LCO-Emb-7B, OmniUE achieves the best performance in 7 of the 8 MAEB meta-task categories, with particularly large gains of +15.1 on Audio Retrieval, +9.5 on Re-ranking, and +4.2 on Multi-label Classification.
Ablation results further show that removing the segmenter streams leads to consistent degradation in performance.
This confirms that segmenter-enhanced representations provide critical fine-grained information for audio-text interaction.
Additional experiments on textual-interactive video-to-audio, audio-to-video, video-audio-to-text, and text-video-audio are included in Appendix \ref{app:additional-textual}.

\subsection{Visual-Interactive TI2T Performance (SCaR)}

\begin{table}
    \centering
    \caption{Results on 5 SCaR tasks. \hlpink{Improvements} are computed relative to the best-performing 2B and 7B baselines. We highlight the \textbf{best} and \underline{second-best} results in each column. Baseline results are from the original SCaR paper.}
    \setlength{\tabcolsep}{15pt} 
    \scalebox{0.65}{
        \begin{tabular}{lcccccc}
    \toprule
     \textbf{Model $\downarrow$ \; $\mid$ \; Datasets $\rightarrow$} & RefCOCO+ & RefCOCOg & VisualGenome & COCO-Stuff & ADE20K & \textbf{Overall} \\
    \midrule
    \multicolumn{7}{c}{\textbf{Textual-Interactive-Only Models}} \\
    \midrule
    VLM2Vec-2B \cite{DBLP:conf/iclr/JiangMYYZC25} & 24.5 & 29.5 & 22.3 & 19.4 & 24.6 & 24.1 \\
    \rowcolor{gray!10}
    VLM2Vec-7B \cite{DBLP:conf/iclr/JiangMYYZC25} & 23.2 & 29.1 & 14.7 & 25.0 & 22.4 & 22.9 \\
    MMRet-7B \cite{DBLP:conf/acl/0001XLLXWZZL25} & 27.1 & 21.8 & 15.2 & 26.0 & 22.6 & 22.5 \\ 
    \rowcolor{gray!10}
    UniME-7B \cite{DBLP:journals/corr/abs-2510-13515} & 31.4 & 32.8 & 19.0 & 25.3 & 23.0 & 26.3 \\
    \midrule
    \multicolumn{7}{c}{\textbf{Textual + Visual-Interactive Models}} \\
    \midrule
    VIRTUE-2B \cite{wang2026virtuevisualinteractivetextimageuniversal} & 28.8 & 42.4 & 24.4 & 29.9 & 27.5 & 30.6 \\
    \rowcolor{gray!10}
    VIRTUE-7B \cite{wang2026virtuevisualinteractivetextimageuniversal} & 33.0 & 35.3 & 19.6 & 27.1 & 23.8 & 27.8 \\
    
    \midrule
    \rowcolor{sonyblue!15}
    \multicolumn{7}{c}{\textbf{Omni-Interactive Models (Ours)}} \\
    \midrule
    OmniUE-3B & \underline{64.4} & \underline{57.5} & \textbf{36.8} & \underline{57.0} & \underline{41.0} & \underline{51.3} \\
    \rowcolor{gray!10}
    OmniUE-7B & \textbf{65.3} & \textbf{66.2} & \underline{35.5} & \textbf{61.6} & \textbf{47.5} & \textbf{55.2} \\
    \rowcolor{pink!30!white}
    Improvements (3B) & +35.6 & +15.1 & +12.4 & +27.1 & +13.5 & +20.7 \\
    \rowcolor{pink!30!white}
    Improvements (7B) & +32.3 & +30.9 & +15.9 & +34.5 & +23.7 & +27.4 \\
    \bottomrule
\end{tabular}

    }
    \label{tab:scar-results}
\end{table}

To evaluate visual-interactive capabilities, we compare OmniUE with state-of-the-art visual-interactive models on the SCaR benchmark, where the goal is to find the most relevant text caption based on a given image, text instruction, and a bounding box.
As shown in Tab.~\ref{tab:scar-results}, OmniUE consistently outperforms all baselines across all five datasets.
Specifically, OmniUE achieves significant improvements of 20.7 and 27.4 points over the strongest 2B and 7B baselines, respectively.
These results demonstrate that OmniUE generalizes beyond textual-interactive scenarios to visual-interactive settings.
In addition to training on omni-modal datasets, we attribute these gains to its unified design, which integrates global context with interaction prompts and leverages context aggregation with learnable tokens to capture fine-grained information.

\subsection{Omni-Interactive TVA2A Performance (OmniCHOIR)}

\begin{table}[t]
    \centering
    \caption{Results on OmniCHOIR. We highlight the \textbf{best} results in each column. Text refers to the interaction text prompt. For span scenarios, we 1) convert the selected span into text using the same instructions as in OmniUE and 2) directly crop the corresponding audio segment, reported as text/crop.}
    \setlength{\tabcolsep}{14pt} 
    \scalebox{0.60}{
        \begin{tabular}{lccccccc}
    \toprule
     \textbf{Model $\downarrow$ \; $\mid$ \; Conditions $\rightarrow$} & Text & Span & Mask & Text+Span & Text+Mask & Span+Mask & Text+Span+Mask \\
    \midrule
    \multicolumn{8}{c}{\textbf{Two-Tower-Based Models}} \\
    \midrule
    laion/larger\_clap\_general \cite{DBLP:conf/icassp/WuCZHBD23} & 16.1 & 1.7/12.7 & - & 12.7 & - & - & - \\

    \midrule
    \multicolumn{8}{c}{\textbf{Textual-Interactive Omni-Modal Models}} \\
    \midrule
    ImageBind \cite{DBLP:conf/cvpr/GirdharELSAJM23} & 16.8 & 3.3/13.5 & - & 18.1 & - & - & - \\
    LCO-Emb-3B \cite{xiao2025scaling} & 19.6 & 18.7 & - & 20.5 & - & - & - \\
    LCO-Emb-7B \cite{xiao2025scaling} & 22.3 & 22.4 & - & 23.8/23.8 & - & - & - \\
    WAVE-7B \cite{tang2026wave} & 20.0 & 23.1/23.4 & - & 24.5/24.7 & - & - & - \\
    
    \midrule
    \rowcolor{sonyblue!15}
    \multicolumn{8}{c}{\textbf{Omni-Interactive Omni-Modal Models (Ours)}} \\
    \midrule
    OmniUE-3B & 23.6 & 25.3 & 25.6 & 23.6 & 25.7 & 25.7 & 26.4 \\
    OmniUE-7B & \textbf{25.5} & \textbf{26.3} & \textbf{26.7} & \textbf{27.4} & \textbf{26.7} & \textbf{27.6} & \textbf{28.4} \\
    \bottomrule
\end{tabular}

    }
    \label{tab:omnichoir-results}
\end{table}

To further evaluate omni-interactive capabilities, we stress-test OmniUE on the proposed OmniCHOIR benchmark, which includes seven uni and multimodal interaction conditions, i.e., text, span, mask, text+span, text+mask, span+mask, text+span+mask.
As OmniCHOIR requires omni-modal inputs, we compare against two-tower-based and textual-interactive omni-modal models that only accept text-based interaction prompts.
As shown in Tab.~\ref{tab:omnichoir-results}, OmniUE already outperforms all baselines under the text-only condition, which we attribute to the audio segmenter’s ability to leverage textual cues to focus on user-specified audio.
In particular, both visual (masks) and audio (temporal span) interactions individually outperform the text-only baselines.
The deleterious performance of converting spans with text or cropped intervals demonstrates that the gains of OmniUE cannot be explained by simple span-cropping or text-conversion heuristics, but arise from the proposed omni-interaction representation.
Moreover, combining multiple interaction modalities leads to additional improvements, and using all available modalities yields the best performance.
This trend highlights a key property of OmniUE: it not only supports heterogeneous interaction modalities, but also effectively composes them to refine retrieval.
These results demonstrate that OmniUE can flexibly leverage both individual and combined user interactions, enabling unified omni-interactive retrieval beyond text-only paradigms.

\vspace{-5pt}
\section{Conclusion}
In this paper, we propose OmniUE, the first omni-interactive universal embedder that jointly encodes omni-modal inputs and omni-modal interactions as holistic context and fine-grained guidance.
Distinct from existing approaches that support only text and image as interaction modalities, OmniUE leverages off-the-shelf SAM-3 and SAM-Audio as the visual and audio segmenters with multi-token layer aggregation, enabling the model to produce user-conditioned query embeddings and unlock emerging capabilities in visual and audio interactive embedding scenarios.
To comprehensively evaluate these capabilities, we introduce OmniCHOIR, an omni-interactive TVA2A retrieval benchmark that incorporates text, visual masks, and audio spans as interaction media, along with challenging negative candidates constructed by altering background sounds or target sound categories.
Extensive experiments across textual, visual, audio, and joint interactive embedding benchmarks demonstrate that OmniUE not only outperforms existing methods on their respective tasks, but also enables new functionalities under diverse combinations of interaction modalities.
We believe OmniUE serves as a generic framework for omni-interactive, omni-modal embeddings, while OmniCHOIR offers a new evaluation paradigm for assessing such capabilities, together paving the way toward universal embedders.
Further discussions on limitations and broader impacts are in Appendix \ref{app:limitation}.


\bibliographystyle{abbrvnat}
\bibliography{neurips_2026}



\newpage
\appendix
\section{Limitation and Broader Impacts}
\label{app:limitation}
\noindent\textbf{Limitations.}
While OmniUE represents a significant step toward reinforcing omni-interactive capabilities in embedding models, the main limitation lies in its dependence on pre-trained models (e.g., Qwen2.5-omni and SAM-Audio) since it is infeasible to train all modules from scratch.
\citet{xiao2025scaling} introduce a generation-representation scaling law showing that embedding performance is positively related to generative performance, suggesting that our proposed method can be boosted by incorporating future state-of-the-art pretrained models.
Nonetheless, this important and orthogonal direction reflects our overarching objective: to pave the way for embedding models that can process any kind of user intent, while maintaining robust universal embedding performance.
In addition, there is a tradeoff between computational efficiency and interaction precision when using masks as interaction media (see Appendix \ref{app:inference-analysis}); however, we provide further analysis using bounding boxes as alternative efficient visual-interactive solutions.
In future work, we aim to explore more use scenarios for OmniUE (e.g., interactive similarity metrics), advance our framework, and build other complementary combinations as interactive benchmarks, to fully unlock the potential of omni-interactive universal embedders.
Moreover, we plan to scale OmniCHOIR to a larger collection.

\noindent\textbf{Broader Impacts.}
As an omni-interactive, omni-modal embedder, OmniUE can facilitate the development of interaction-aware models (e.g., \cite{cong2025viva}) by providing effective user-conditioned embeddings.
In addition, OmniUE can serve as a similarity-based evaluation model, similar to prior approaches such as ImageBind~\cite{DBLP:conf/cvpr/GirdharELSAJM23}.
However, unlike conventional methods that rely on holistic similarity, OmniUE enables interaction-aware similarity by jointly modeling multimodal inputs and user-provided interaction prompts.
Finally, OmniUE can be integrated into agentic pipelines that rely on contextual embeddings for applications such as memory, personalization, and adaptive decision-making, where its interaction-aware representations enable more fine-grained and controllable behavior.
As a representation learning framework, OmniUE poses a lower direct risk of generating harmful or deceptive content compared to generative multimodal models. 
However, its fine-grained interaction-aware embeddings could potentially be misused in downstream applications. For instance, the integration of multimodal user signals necessitates careful attention to data privacy to prevent the unintended encoding of sensitive personal attributes.
\section{LLM Usage}
\label{app:llm-usage}
In addition to Qwen3-Omni used for building the OmniCHOIR benchmark (i.e., sound category selection), we use LLMs solely for polishing the manuscript.

\begin{table}[h]
    \centering
    \caption{Detailed configurations for OmniUE 3B and 7B models.}
    \setlength{\tabcolsep}{10pt}
    \scalebox{0.9}{
        \begin{tabular}{lcc}
    \toprule
     Configurations & OmniUE-3B & OmniUE-7B \\
    \midrule \midrule
    Omni-LLM & Qwen2.5-Omni-3B & Qwen2.5-Omni-7B \\
    \rowcolor{gray!10}
    Vision Segmenter & \multicolumn{2}{c}{facebook/sam3} \\
    Audio Segmenter & \multicolumn{2}{c}{facebook/sam-audio-base} \\
    \rowcolor{gray!10}
    \# of heads in $\text{SAC}$ ($h$) & \multicolumn{2}{c}{2} \\
    $|F^H_{vs}| \times |F^W_{vs}|$ & \multicolumn{2}{c}{288$\times$288} \\
    \rowcolor{gray!10}
    $|F_{as}|$ & \multicolumn{2}{c}{maximum 250} \\
    $|F'_{vs}|$ & \multicolumn{2}{c}{maximum 256} \\
    \rowcolor{gray!10}
    $|F'_{as}|$ & \multicolumn{2}{c}{maximum 128} \\
    $d_{vs}$ & \multicolumn{2}{c}{256} \\
    \rowcolor{gray!10}
    $d_{as}$ & \multicolumn{2}{c}{128} \\
    Max sequence length & \multicolumn{2}{c}{20480} \\
    \rowcolor{gray!10}
    Max frame resolution & \multicolumn{2}{c}{156800} \\
    Iterative sampling procedure & \multicolumn{2}{c}{8-step ODE solver} \\
    \rowcolor{gray!10}
    A\textsubscript{S}-L connector \& V\textsubscript{S}-L connector & \multicolumn{2}{c}{Random initialized} \\
    Audio sampling rate & \multicolumn{2}{c}{16kHz} \\
    \rowcolor{gray!10}
    LoRA rank & \multicolumn{2}{c}{32} \\
    LoRA dropout & \multicolumn{2}{c}{0.1} \\
    \rowcolor{gray!10}
    LoRA alpha & \multicolumn{2}{c}{64} \\
    Temperature $\tau$ & \multicolumn{2}{c}{0.02} \\
    \rowcolor{gray!10}
    Batch size & \multicolumn{2}{c}{1024} \\
    Training steps & \multicolumn{2}{c}{3300} \\
    \rowcolor{gray!10}
    Warmup steps & \multicolumn{2}{c}{200} \\
    Learning rate & \multicolumn{2}{c}{2e-5} \\
    \rowcolor{gray!10}
    $d$ & 2048 & 3584 \\
    $l$ & \multicolumn{2}{c}{4} \\
    \rowcolor{gray!10}
    $K$ & \multicolumn{2}{c}{4} \\
    GPU & \multicolumn{2}{c}{8$\times$H100 80G} \\
    \rowcolor{gray!10}
    Precision & \multicolumn{2}{c}{bf16} \\
    Optimizer & \multicolumn{2}{c}{AdamW ($\beta_1=0.9, \beta_2=0.999$)} \\
    \rowcolor{gray!10}
    Training time & 177 hours & 195 hours \\
    \bottomrule
\end{tabular}

    }
    \label{tab:implementation-details}
\end{table}

\section{Implementation Details}
\label{app:implementation-details}

\subsection{Parameter Setting}

OmniUE-3B and OmniUE-7B are trained with a $\tau$ of 0.02 and a learning rate of 2e-5.
The OmniUE 3B and 7B training was conducted on 8$\times$H100 80GB, which took around 177 and 195 hours, respectively.
The remaining configurations are identical to the default settings of Qwen2.5-omni, SAM-Audio, and SAM-3.
The iterative sampling procedure is the ODE solver following the original SAM-Audio paper.
While SAM-Audio originally uses 16 steps, we found it performs on par with 8 steps, while reducing GPU memory usage.
The kernel size and stride of Conv2D and Conv1D are set to 4.
Detailed configurations are summarized in Tab. \ref{tab:implementation-details}.

\subsection{Training Datasets}
\begin{table}[h]
    \centering
    \caption{Overview of the training data used to train OmniUE. * denotes synthesized text captions.}
    \setlength{\tabcolsep}{10pt}
    \scalebox{0.9}{
        \begin{tabular}{lcc}
    \toprule
    Data Source & Modalities & \# of Samples \\
    \midrule \midrule
    AudioSet \cite{DBLP:conf/icassp/GemmekeEFJLMPR17} & Video, Audio, Text* & 86k \\
    \rowcolor{gray!10}
    WavCaps \cite{DBLP:journals/taslp/MeiMLKKZPZW24} & Audio, Text & 404k \\
    MSR-VTT \cite{DBLP:conf/cvpr/XuMYR16} & Video, Text & 180k \\
    \rowcolor{gray!10}
    PE-Videos \cite{DBLP:journals/corr/abs-2504-13181} & Video, Text & 983k \\
    VALOR \cite{DBLP:journals/pami/LiuCHGZWT25} & Video, Audio, Text & 2k \\
    \rowcolor{gray!10}
    VATEX \cite{DBLP:conf/iccv/WangWCLWW19} & Video, Text & 5k \\
    AudioCaps \cite{DBLP:conf/naacl/KimKLK19} & Audio, Text & 45k \\
    \rowcolor{gray!10}
    Clotho \cite{DBLP:conf/icassp/DrossosLV20} & Audio, Text & 19k \\
    Panda70m \cite{DBLP:conf/cvpr/ChenSMDCJF0RYT24} & Video, Audio, Text* & 278k \\
    \rowcolor{gray!10}
    Shot2Story \cite{DBLP:conf/iclr/0002YC0W25} & Video, Audio, Text & 970k \\
    VGGSound \cite{DBLP:conf/icassp/ChenXVZ20} & Video, Audio, Text* & 183k \\
    \rowcolor{gray!10}
    Youtube8m \cite{DBLP:journals/corr/Abu-El-HaijaKLN16} & Video, Audio, Text* & 269k \\
    \bottomrule
\end{tabular}
    }
    \label{tab:training-data}
\end{table}

Tab.~\ref{tab:training-data} summarizes the data sources used and their corresponding modalities, resulting in a total of approximately 3.5M multimodal samples.
To enrich modality coverage, we use Qwen/Qwen3-Omni-30B-A3B-Instruct~\cite{Qwen3-Omni} to synthesize text captions from paired video and audio inputs, where the generated captions jointly capture both visual and auditory information.
For VGGSound, we adopt text captions from AudioSetCaps~\cite{bai2024audiosetcapsnipsws}.
Our early experiments suggest that incorporating these synthesized captions improves embedding performance on MMEB-v2-video and MAEB.
It is important to note that the training datasets used in OmniUE do not overlap with the OmniCHOIR samples.
\section{OmniCHOIR Details}
\label{app:omnichoir-details}

\subsection{OmniCHOIR Examples}
For better readability, we present OmniCHOIR examples in the supplementary material (i.e., index.html), where it can be seen that the benchmark cover diverse domains.

\subsection{Prompt Template for Video-Audio Captioning}
\label{app:scar-build-template}
We summarize the prompt template for Qwen3-Omni-30B below.

\begin{tcolorbox}[title=Prompt Template for Qwen3-Omni, myprompt]
\scriptsize
You are an expert audio-visual scene analyst. Given a video and its primary audio, infer one background sound that naturally fits the visible scene, and generate several distractor background sounds that are plausible in the same broad environment but are not supported by the video and are semantically different from the gold sound. Follow the scene strictly: do not invent unlikely entities or actions that are not visible or strongly implied.

The primary sound in this video is: "{original\_caption}".

Below is the complete list of allowed environmental sound categories. You MUST choose all outputs ONLY from this list, using exact string matches:{esc50\_categories}

\textbf{Task:}

1. Select exactly ONE category from the allowed list that would plausibly occur as a background sound in this video's scene.

2. Select exactly FIVE other categories from the same allowed list that are plausible in the broad environment, but are not strongly supported by the video context and are semantically different from the gold category.

3. Be creative and diverse negative categories.

    \textbf{Strict constraints:}
    
    - Every output must be an exact category name copied from the allowed list.
    
    - Do NOT output any sound label outside the list, even if it seems more natural.
    
    - Do NOT output the primary sound itself.
    
    - Do NOT output synonyms, paraphrases, or near-duplicates of a listed category.
    
    - DO NOT use music\_playing that is not in the list, use the closest plausible allowed background category instead.
    
    - Negative backgrounds must be realistic distractors in the same broad scene, not absurdly impossible sounds.
    
    - Negative backgrounds must be different from the gold category and from each other.
    
Respond ONLY with valid JSON (no extra text):
\begin{verbatim}
{
    "gold_background": "<exact category from list>",
    "negative_backgrounds": 
        [
            "<exact category from list>", 
            "<exact category from list>", 
            "<exact category from list>", 
            "<exact category from list>",
            "<exact category from list>"
        ]
}
\end{verbatim}
\end{tcolorbox}
\section{Additional Experiments}

\subsection{Ablation and Parameter Study}
\label{app:ablation-study}
To analyze the relative contributions of different components and parameters, we conduct ablation and parameter studies on 15 variant designs and choices.
To iteratively refine the results, all studies here are experimented with a batch size of 512 and 3000 training steps on OmniUE-3B, which take around 48 hours for each training configuration.
The evaluation on OmniCHOIR uses text as conditions.

\begin{table}[h]
    \centering
    \caption{Ablation study on the impact of each proposed component. "w/o visual and audio streams" denotes the removal of visual and audio interactions during both training and testing stages. "w/o visual stream" and "w/o audio stream" denote the removal of visual-only and audio-only interactions during both stages. "w/o multi-token inputs" replaces the $K$ learnable tokens with the EOS hidden state, reducing $W$ to $\mathbb{R}^{M}$. "w/o layer aggregation" removes the context aggregation module and applies mean pooling over the $K$ tokens from the last layer.}
    \begin{tabular}{lcccc}
    \toprule
     & MMEB-v2-video & MAEB & SCaR & OmniCHOIR \\
    \midrule
    \rowcolor{sonyblue!15}
    OmniUE-3B & 44.7 & 47.8 & 50.5 & 22.2 \\
    w/o visual stream & 42.7 & 47.6 & 32.0 & 20.5 \\
    w/o audio stream & 44.1 & 47.4 & 49.8 & 18.7 \\
    w/o visual and audio streams & 42.5 & 47.5 & 30.9 & 16.3 \\
    w/o multi-token inputs & 43.8 & 46.6 & 47.2 & 20.1 \\
    w/o layer aggregation & 43.2 & 45.0 & 49.8 & 18.5 \\
    \bottomrule
\end{tabular}

    \label{tab:ablation-removal}
\end{table}

\noindent\textbf{Relative contributions of each module.}
As shown in Tab.~\ref{tab:ablation-removal}, removing any proposed component consistently degrades performance across all interaction benchmarks, demonstrating that each module provides complementary benefits.
In particular, removing the visual or audio interaction streams leads to substantial drops on SCaR and OmniCHOIR, confirming that these streams are critical for modality-specific interactions.
Furthermore, removing the visual stream negatively impacts MMEB-v2-video and SCaR, while removing the audio stream hurts more on MAEB; both removal settings are inferior to OmniCHOIR compared to OmniUE.
Overall, the results establish that 1) multi-token inputs and 2) layer aggregation yield consistent gains across all benchmarks, while 3) interaction streams provide targeted improvements in their corresponding visual and audio scenarios.

\begin{figure}[h]
    \centering
    \includegraphics[width=\textwidth]{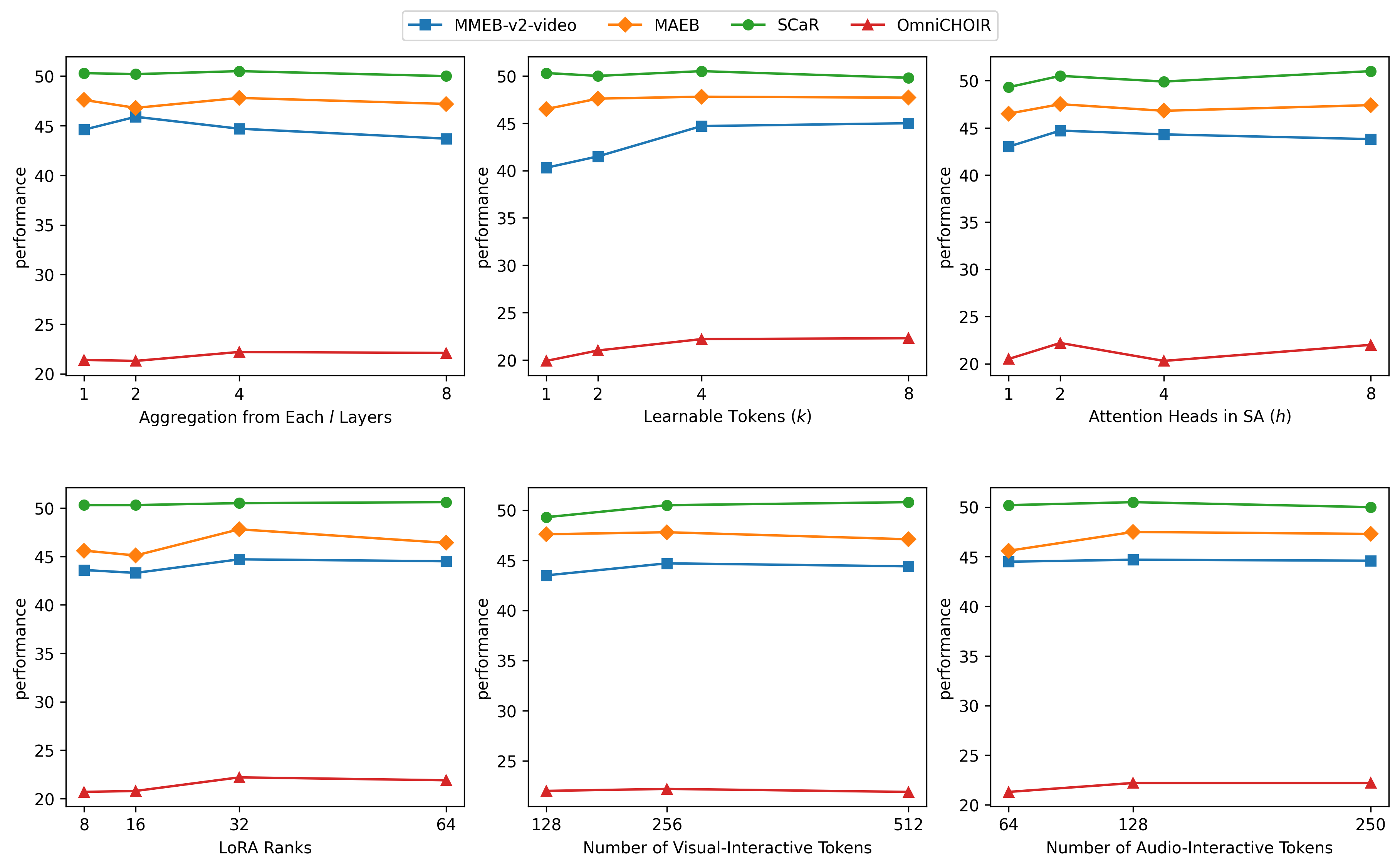}
    \vspace{-20pt}
    \caption{
    Parameter study on 1) Aggregation from each $l$ layers; 2) \# of learnable tokens ($k$); 3) \# of attention heads in SA ($h$); 4) LoRA ranks; 5) number of visual-interactive tokens ($|F'_{vs}|$); and 6) number of audio-interactive tokens ($|F'_{as}|$).}
    \label{fig:parameter-study}
\end{figure}

\noindent\textbf{Relative contributions of omni-interactions and architectural changes.}
To further isolate the contribution of our proposed components, we additionally perform two complementary full-training ablations: 1) removing the proposed multi-token inputs and layer aggregation (architectural refinements), and 2) removing the proposed visual and audio interaction streams (omni-interaction formulation).
As shown in Tab. \ref{tab:interaction-architecture}, removing the interaction streams causes the largest degradation on interaction-centric benchmarks (SCaR: -18.4, OmniCHOIR: -7.1), demonstrating that the proposed omni-interaction formulation is the primary factor enabling new interaction capabilities. Meanwhile, removing multi-token inputs and layer aggregation consistently reduces performance across benchmarks, confirming that these architectural designs further improve embedding quality.
Together, these results suggest that OmniUE's improvements arise from the proposed omni-interaction formulation and embedding design built on top of pretrained models, rather than solely from the pretrained modules.

\begin{table}[h]
    \centering
    \caption{Ablation study on the omni-interactions and architectural changes.}
    \scalebox{0.9}{
        \begin{tabular}{lcccc}
    \toprule
     & MMEB-v2-video & MAEB & SCaR & OmniCHOIR \\
    \midrule
    \rowcolor{sonyblue!15}
    OmniUE-3B & 48.4 & 50.5 & 51.3 & 26.4 \\
    w/o learnable tokens \& context aggregation & 43.4 & 44.7 & 46.3 & 23.5 \\
    w/o visual \& audio streams & 46.5 & 48.2 & 32.9 & 19.3 \\
    \bottomrule
\end{tabular}

    }
    \label{tab:interaction-architecture}
\end{table}

\noindent\textbf{Impact of key hyperparameters.}
To evaluate the robustness of OmniUE, we perform a parameter study over six factors: 1) the aggregation interval $l$ across LLM layers, 2) the number of learnable tokens $K$, 3) the number of attention heads in the self-attention (SA) module, 4) different LoRA ranks, 5) the number of visual-interactive tokens ($|F'_{vs}|$), and 6) the number of audio-interactive tokens ($|F'_{as}|$).
As shown in Fig.~\ref{fig:parameter-study}, aggregating more intermediate hidden states generally improves performance compared to sparse aggregation (e.g., $l=8$), with the best overall performance achieved at $l=4$, indicating a balance between information richness and redundancy.
For learnable tokens, using a single token ($K=1$) significantly degrades performance, whereas increasing to $K=2$ consistently improves results, suggesting that multiple tokens are necessary to capture heterogeneous multimodal interactions. 
In addition, using 2 and 8 attention heads yields comparable performance; we adopt 2 heads as the default configuration due to its more favorable performance–efficiency trade-off.
Finally, the ablations on LoRA rank, visual-interactive tokens, and audio-interactive tokens further validate our design choices.
Varying the number of visual-interactive tokens primarily affects performance on visual benchmarks, while varying the number of audio-interactive tokens predominantly impacts audio benchmarks.
This modality-aligned sensitivity confirms that the proposed token design effectively captures modality-specific interactions and that our selected configurations achieve a balanced and robust performance across tasks.

\begin{table}[h]
    \centering
    \caption{Ablation study with OmniUE-3B on 1) audio segmenter size, 2) visual segmenter version, 3) segmenter connector design, and 4) layer aggregation variants. The \hlsonyblue{highlighted row} indicates the configurations used in OmniUE-3B and OmniUE-7B.}
    \begin{minipage}[t]{0.49\textwidth}
        \centering
        \scalebox{0.70}{
            \small
            \begin{tabular}{lcccc}
    \toprule
     Choice & MMEB-v2-video & MAEB & SCaR & OmniCHOIR \\
    \midrule
    small & 43.4 & 46.7 & 49.3 & 21.5 \\
    \rowcolor{sonyblue!15}
    base & 44.7 & 47.8 & 50.5 & 22.2 \\
    large & 44.2 & 48.1 & 50.2 & 20.8 \\
    \bottomrule
\end{tabular}

        }
        \label{tab:ablation-sam}
    \end{minipage}
    \begin{minipage}[t]{0.49\textwidth}
        \centering
        \scalebox{0.70}{
            \small
            \begin{tabular}{lcccc}
    \toprule
     Choice & MMEB-v2-video & MAEB & SCaR & OmniCHOIR \\
    \midrule
    SAM-2.1-B+ & 43.9 & 44.5 & 46.1 & 20.7 \\
    \rowcolor{sonyblue!15}
    SAM-3 & 44.7 & 47.8 & 50.5 & 22.2 \\
    \bottomrule
\end{tabular}

        }
        \label{tab:ablation-sam-version}
    \end{minipage}
    \vspace{0.6em}
    
    \begin{minipage}[t]{0.49\textwidth}
        \centering
        \scalebox{0.70}{
            \small
            \begin{tabular}{lcccc}
    \toprule
     Alternative & MMEB-v2-video & MAEB & SCaR & OmniCHOIR \\
    \midrule
    w/o SA & 42.8 & 45.1 & 47.9 & 20.5 \\
    \rowcolor{sonyblue!15}
    w/ SA & 44.7 & 47.8 & 50.5 & 22.2 \\
    \bottomrule
\end{tabular}

        }
        \label{tab:ablation-connector}
    \end{minipage}
    \begin{minipage}[t]{0.49\textwidth}
        \centering
        \scalebox{0.70}{
            \small
            \begin{tabular}{lcccc}
    \toprule
     Alternative & MMEB-v2-video & MAEB & SCaR & OmniCHOIR \\
    \midrule
    MLP & 41.8 & 47.0 & 45.9 & 20.2 \\
    \rowcolor{sonyblue!15}
    Layer-token weights & 44.7 & 47.8 & 50.5 & 22.2 \\
    \bottomrule
\end{tabular}

        }
        \label{tab:ablation-layer-aggregation}
    \end{minipage}
    
    \label{tab:ablation}
\end{table}

\noindent\textbf{Design choices for variant modules.}
To further analyze key design decisions, Tab.~\ref{tab:ablation} compares several alternatives: 1) the pretrained audio segmenter, 2) the pretrained visual segmenter, 3) the role of self-attention (SA) in segmenter-related connectors, and 4) different aggregation strategies.
For segmenters, the base variant of SAM-Audio consistently achieves the overall best performance among available sizes, and SAM-3 outperforms SAM2.1~\cite{DBLP:conf/iclr/RaviGHHR0KRRGMP25} as visual segmenters.
This trend indicates that stronger pretrained segmenters directly translate to improved embedding quality in multimodal interaction tasks.
Removing self-attention (SA) from the connectors leads to consistent performance degradation, demonstrating its importance. This suggests that SA enables the model to selectively attend to different segmentation tokens conditioned on user intent, thereby improving interaction modeling.
Finally, comparing aggregation methods shows that a standard MLP underperforms our proposed layer-token importance weighting. We attribute this to the substantially larger number of trainable parameters in MLP-based aggregation, which introduces optimization bias and reduces generalization, whereas our design provides a more structured and efficient aggregation mechanism.

\subsection{Analysis on Latency and Memory}
\label{app:inference-analysis}

\begin{table}[h]
    \centering
    \caption{Analysis on inference time and memory for unimodal and multimodal query modes of OmniUE-7B on OmniCHOIR.}
    \scalebox{0.93}{
        \begin{tabular}{llccc}
    \toprule
     & Query Mode & Time (per sample) & Memory & Performance \\
    \midrule\midrule
    \multicolumn{5}{c}{\textbf{OmniCHOIR}} \\
    \midrule
    LCO-Emb-7B & Text & 1.0s & 17.7GB & 22.3 \\
    WAVE-7B & Text & 1.3s & 18.2GB & 20.0 \\
    
    \midrule
    OmniUE-7B (Ours) & Text & 1.3s & 20.8GB & 25.5 \\
     & Span & 1.3s & 20.8GB & 26.3 \\
     & Mask & 9.4s & 20.8GB & 26.7 \\
     & Text+Span & 1.3s & 20.8GB & 27.4 \\
     & Text+Mask & 9.3s & 20.8GB & 26.7 \\
     & Span+Mask & 9.4s & 20.8GB & 27.6 \\
     & Text+Span+Mask & 10.5s & 20.8GB & 28.4 \\
     \hdashline
     & Bbox & 1.5s & 20.8GB & 26.4 \\
     & Text+Bbox & 1.5s & 20.8GB & 26.7 \\
     & Text+Span+Bbox & 2.1s & 20.8GB & 27.0 \\
    \bottomrule
\end{tabular}

    }
    \label{tab:latency-memory}
\end{table}

We report the average end-to-end inference latency (s) and GPU memory consumption (GB) across all seven operational modes of OmniUE-7B and LCO-Emb-7B on the OmniCHOIR benchmark.
As illustrated in Tab.~\ref{tab:latency-memory}, while OmniUE introduces a marginal overhead to facilitate omni-interactive capabilities, it achieves a substantial 14.3\% performance gain under text-conditioned settings.
To further optimize the inference pipeline, we investigate the trade-off between segmentation masks and bounding boxes (Bbox).
While substituting masks with Bboxes results in a slight performance penalty, it significantly reduces latency, which provides a practical recipe for balancing computational throughput with high-fidelity performance in real-world applications.

\subsection{Robustness of OmniUE to Interaction Prompts}
To evaluate the robustness of OmniUE to interactions, we conducted additional experiments to analyze the effects of imperfect interaction inputs, including 1) randomly sampled visual masks, 2) randomly sampled audio spans, and 3) removing the text prompt completely.

\begin{table}[h]
    \centering
    \caption{Analysis on noisy masks, spans, and text interaction prompts.}
    \setlength{\tabcolsep}{14pt} 
    \scalebox{0.65}{
    \begin{tabular}{lccccccc}
    \toprule
     \textbf{Model $\downarrow$ \; $\mid$ \; Conditions $\rightarrow$} & Text & Span & Mask & Text+Span & Text+Mask & Span+Mask & Text+Span+Mask \\
    \midrule
    
    \rowcolor{sonyblue!15}
    OmniUE-7B & \textbf{25.5} & \textbf{26.3} & \textbf{26.7} & \textbf{27.4} & \textbf{26.7} & \textbf{27.6} & \textbf{28.4} \\
    \midrule
    1) Random masks & - & - & 16.8 & - & 21.3 & 22.5 & 22.9 \\
    2) Random spans & - & 19.6 & - & 19.8 & - & 20.5 & 23.7 \\
    3) w/o text prompts & 10.7 & - & - & 16.9 & 21.0 & - & 22.3 \\
    \bottomrule
\end{tabular}

    }
    \label{tab:robustness-interaction}
\end{table}

\subsection{Additional Modality Directions with Textual Interactions}
\label{app:additional-textual}
To further support the universal embedder claim, we additionally evaluate video-to-audio (V2A) and audio-to-video (A2V) retrieval on VGGSound, and video-audio-to-text (VA2T) and text-to-video-audio on VALOR-32K, using the standard Recall@1 metric. We compare against the strongest method applicable to each benchmark (WAVE-7B \cite{tang2026wave} on VGGSound and VALOR \cite{DBLP:journals/pami/LiuCHGZWT25} on VALOR-32K). As shown in Tab. \ref{tab:additional-modality}, the results show that OmniUE-7B outperforms the state-of-the-art methods on both benchmarks, further strengthening the capability toward any-to-any retrieval.

\begin{table}[h]
    \centering
    \caption{Performance on V2A/A2V on VGGSound and VA2T/T2VA on VALOR-32K.}
    \begin{minipage}[t]{0.49\textwidth}
        \centering
        \scalebox{1}{
            \begin{tabular}{lcc}
    \toprule
      & V2A & A2V \\
    \midrule
    WAVE-7B & 25.0 & 25.8 \\
    \rowcolor{sonyblue!15}
    OmniUE-7B & 27.2 & 26.1 \\
    \bottomrule
\end{tabular}

        }
        \label{tab:vggsound}
    \end{minipage}
    \begin{minipage}[t]{0.49\textwidth}
        \centering
        \scalebox{1}{
            \begin{tabular}{lcc}
    \toprule
      & VA2T & T2VA \\
    \midrule
    VALOR & 74.5 & 73.2 \\
    \rowcolor{sonyblue!15}
    OmniUE-7B & 78.9 & 79.3 \\
    \bottomrule
\end{tabular}

        }
        \label{tab:valor}
    \end{minipage}
    
    \label{tab:additional-modality}
\end{table}

\end{document}